\documentclass{article}
\usepackage{spconf,amsmath,graphicx}

\usepackage{times}
\usepackage{soul}
\usepackage{url}

\usepackage{comment}
\usepackage{amssymb} 
\usepackage{color, colortbl}

\usepackage{xcolor}
\usepackage{nicefrac}
\usepackage{booktabs}
\usepackage{placeins}
\usepackage{pifont}
\usepackage{subcaption}
\usepackage{arydshln}
\usepackage{adjustbox}

\usepackage{listings}
\usepackage{multirow}

\newcommand{\padd}[1]{\small{\textcolor{black}{($\uparrow$#1)}}}

\usepackage[utf8]{inputenc}

\usepackage{graphicx}
\usepackage{amsthm}
\usepackage{algorithm}
\usepackage{algorithmic}
\usepackage[switch]{lineno}

\title{Generalizable Two-Branch Framework for Image Class-Incremental Learning}
\name{Chao Wu$^{1}$, Xiaobin Chang$^{1,*}$\thanks{$^*$ indicates corresponding author.}, Ruixuan Wang$^{2}$}
\address{ $^1$School of Artificial Intelligence, Sun Yat-sen University, China \\
$^2$School of Computer Science and Engineering, Sun Yat-sen University, China\\
}
\begin{document}
%
\maketitle
\begin{abstract}
 Deep neural networks often severely forget previously learned knowledge when learning new knowledge. Various continual learning (CL) methods have been proposed to handle such a catastrophic forgetting issue from different perspectives and achieved substantial improvements.
    In this paper, a novel two-branch continual learning framework is proposed to further enhance most existing CL methods. Specifically, the main branch can be any existing CL model and the newly introduced side branch is a lightweight convolutional network. The output of each main branch block is modulated by the output of the corresponding side branch block. Such a simple two-branch model can then be easily implemented and learned with the vanilla optimization setting without whistles and bells.
    Extensive experiments with various settings on multiple image datasets show that the proposed framework yields consistent improvements over state-of-the-art methods.
\end{abstract}
\begin{keywords}
{\color{black}Incremental Learning, Neural Networks, Image classification}
\end{keywords}


\section{Introduction}
\begin{figure}[t]
	\centering

    \includegraphics[width=0.95\linewidth]{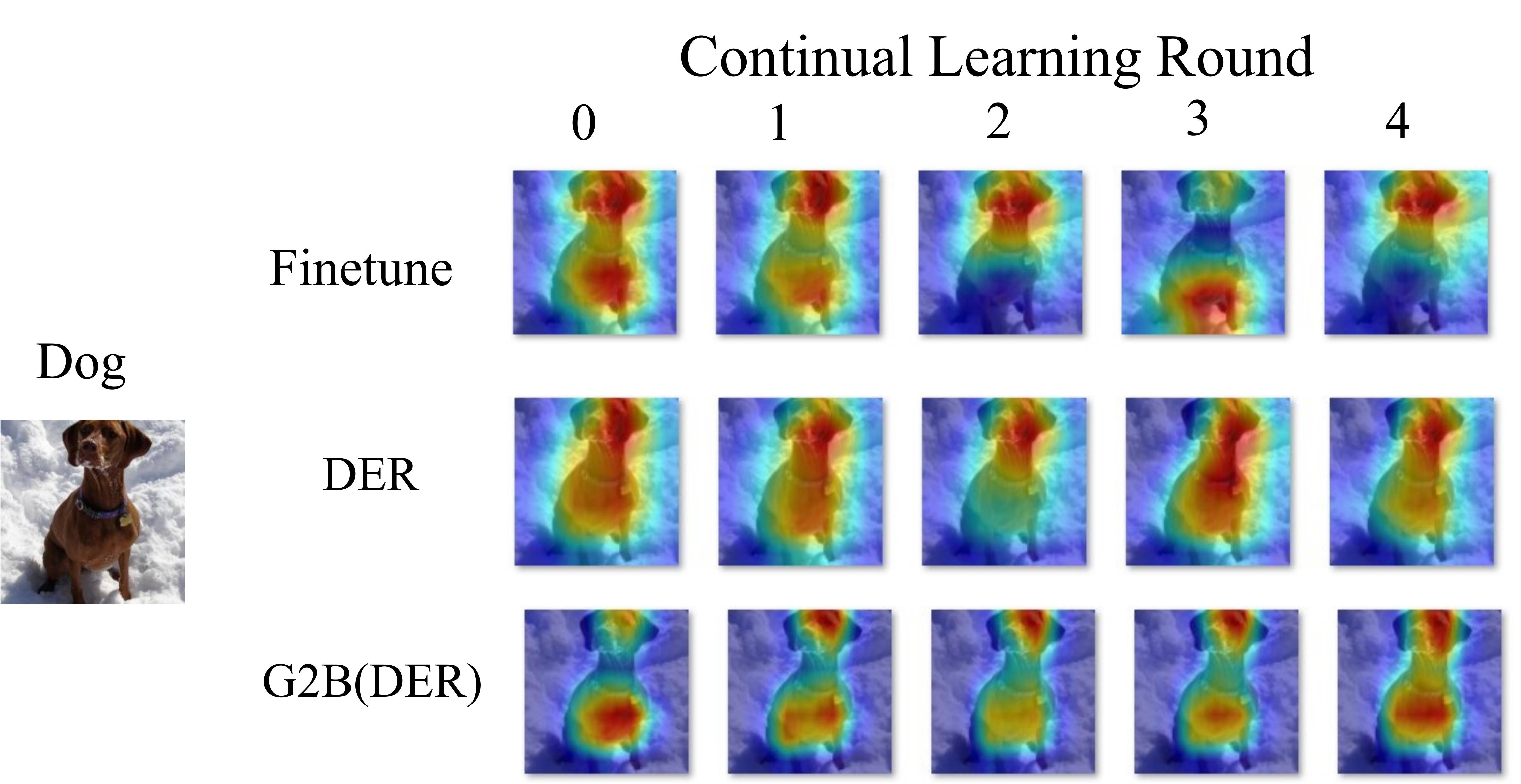}
	\caption{
 The feature attention maps from different methods are illustrated. The `Dog' class only appears in round 0 during continual learning. The models are then continually updated with other classes at different rounds. `Finetune': model is simply finetuned across rounds. `DER': a CL method. `G2B(DER)': our G2B framework with DER.
 }
	\label{fig:intro_att}
\end{figure}

One exceptional ability of humans is to continually learn and accumulate knowledge over time. However, current deep neural network models would catastrophically forget previously learned knowledge when learning new knowledge~\cite{french1999catastrophic,McCloskey1989CatastrophicII}.
Since knowledge is implicitly stored in model parameters, the change in model parameters during continual learning of new knowledge would inevitably deteriorate model memory of old knowledge. 
Taking the class-incremental learning (CIL) setting as an example, the attention of old task image features extracted by the network learned on the new tasks drifts away from the discriminative areas (Figure~\ref{fig:intro_att}, first row), resulting in inferior performance.

To alleviate the catastrophic forgetting issue, multiple approaches have been proposed to help deep learning models achieve the continual learning ability, either by trying to keep model parameters or model responses relevant to old knowledge from changing~\cite{douillard2020podnet,li2017lwf,kirkpatrick2017ewc,wu2019bias_correction}, rehearsing old knowledge with a preserved small subset of old data~\cite{Buzzega2020DarkEF,Benjamin2018MeasuringAR,Ars2018agem,rebuffi2017icarl}, or dynamically growing the model components specifically for new knowledge~\cite{DyTox,yan2021der,yoon2017lifelong,zhang2020side}. 
Jointly applying the rehearsal and dynamic growth strategies, e.g, DyTox~\cite{DyTox} and DER~\cite{yan2021der}, have achieved state-of-the-art performance in class-incremental learning of image classes.
As shown in Figure~\ref{fig:intro_att} (second row), the catastrophic forgetting issue can be alleviated by \textcolor{black}{DER} as its feature attention areas are more discriminative and consistent than the baseline `Finetune' across rounds of learning.
In this paper, a simple yet effective generalizable two-branch (G2B) framework is proposed to enhance the performance of existing strategies.
Specifically, the main branch can be any existing continual learning strategy based on a convolutional neural network (CNN) or vision transformer (ViT), and the lightweight side branch is an independent convolutional network.
The side branch interacts with the main branch such that the output of each block in the main branch is modulated by the output of the corresponding block in the side branch, resulting in relatively sparse activations as the outputs of the main branch blocks.
With such sparse activated outputs, the proposed method can be more resistant to the forgetting problem in continual learning.
As illustrated in Figure~\ref{fig:intro_att} (last row), the attention maps of the sparsely activated features of our method are kept compact and focus on the discriminative visual areas. 
Training the G2B framework is painless as the vanilla optimization setting can be applied directly. Extensive empirical evaluations on multiple image datasets show that the G2B framework yields consistent improvements over state-of-the-art methods.

\section{Methodology}\label{sec:method}
In class-incremental learning (CIL), the model is trained from an ordered sequence of datasets. 
When the model is updated to learn a set of new classes with associated training data each time, it assumes that there is no or only very few old data available for each of the previously learned (old) classes. Thus, changes in model parameters during continual learning would make the updated model have seriously biased predictions toward recently learned new classes. In other words, some model parameters which are crucial to the knowledge preservation of old classes may be changed during learning new classes, causing the updated model to catastrophically forget at least part of old knowledge. 

To reduce the possibility of changing model parameters relevant to old knowledge, one way is to update only part of model parameters
(e.g., kernels in CNN classifier) during learning new classes of knowledge. With this consideration, we propose a two-branch framework where the side branch modulates the activation of the main branch at each block, such that a relatively smaller number of model components in the main branch are updated during continual learning.


\subsection{The two-branch framework}

\begin{figure}[t]
	\centering
	\includegraphics[width=0.5\textwidth]{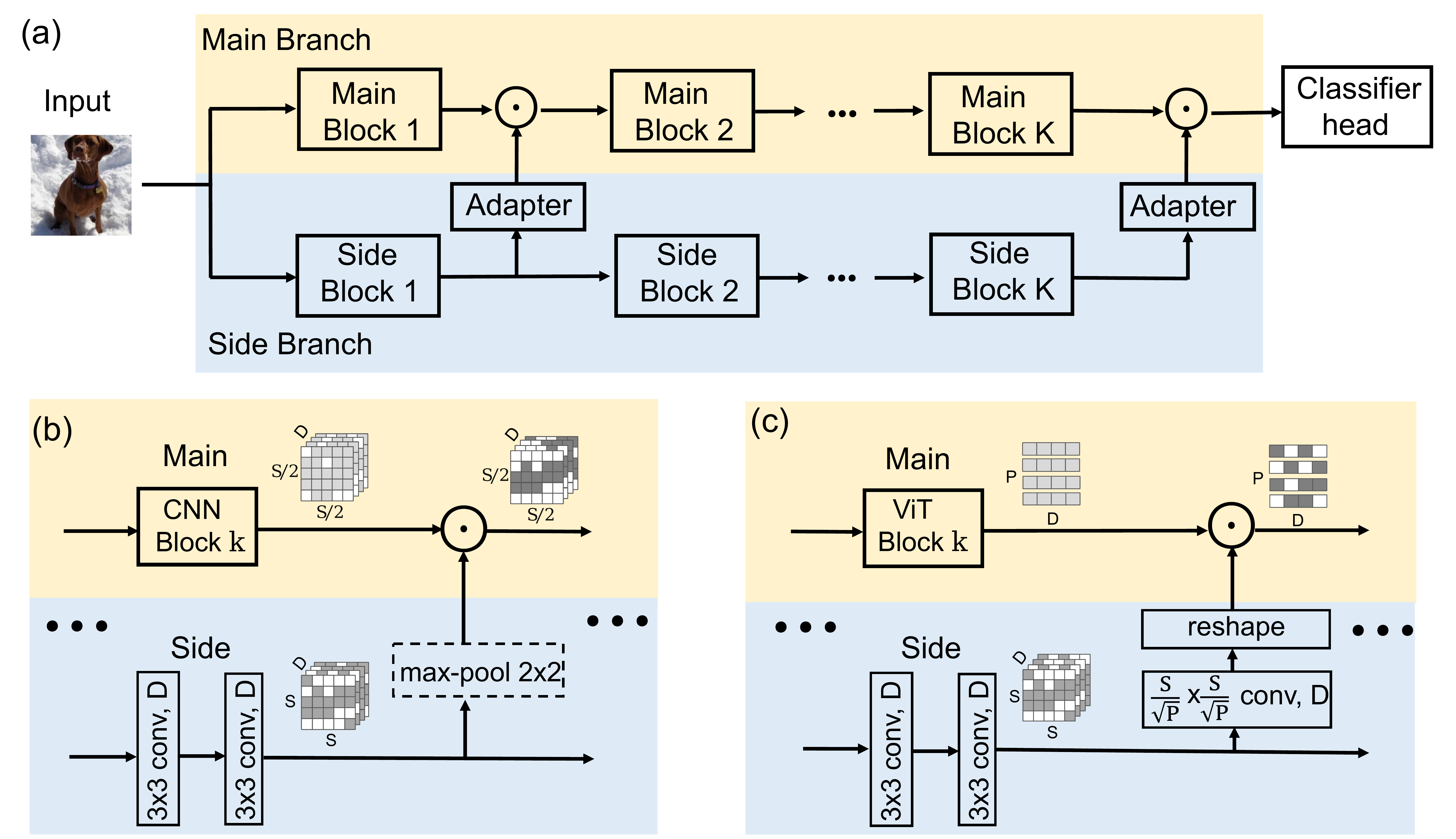}
	\caption{
 {\color{black} The proposed two-branch framework for continual learning.  
 The main branch (shown in yellow background) can be any 
 continual learning framework with either CNN or Transformer backbone, and the side branch (shown in blue background) is a lightweight CNN. (a) The output of a main block is modulated by the corresponding side block.
 (b-c) Design of each side block and the adapter respectively for the CNN backbone and the ViT backbone of the main branch.
 Note that the BN and ReLU operators following the convolutional operator in each side block are omitted for simplicity.  
 }
     }
	\label{fig:framework}
\end{figure}

The proposed framework consists of two branches, as illustrated in Figure~\ref{fig:framework}(a). The main branch can be any existing deep learning framework for continual learning, e.g., iCaRL and DER with a CNN backbone or DyTox with a Transformer backbone.
The side branch is a lightweight convolutional network which has the same number of blocks as the main branch 
and each block consists of only two convolutional layers, as depicted in Figure~\ref{fig:framework} (b) and (c). Therefore, it can be considered as a \emph{plug-in} module for any existing continual learning framework. The side branch interacts with the main branch at each block, using the output of the side branch as a soft mask to modulate the output of the main branch at each block. Formally,
given any input image $\mathbf{x}$, denote by $\mathbf{F}_i(\mathbf{x})$ and $\mathbf{M}_i(\mathbf{x})$ the original outputs of the $i$-block respectively from the main branch and the side branch. 
Then, the modulated output $\hat{\mathbf{F}}_i$ for the $i$-block of the main branch can be obtained by
\begin{equation} \label{eq:modulation}
	\hat{\mathbf{F}}_i(\mathbf{x}) = \mathbf{F}_i(\mathbf{x}) \odot g(\mathbf{M}_i(\mathbf{x})) \,,
\end{equation}
where $\odot$ is the Hadamard product operator, and $g(\cdot)$ is an adapter function particularly for heterogeneous model backbones between the two branches.

When the model backbone of the main branch is a CNN, the number of kernels in each side block is designed to be consistent with those of the corresponding main block, as shown in Figure~\ref{fig:framework}\textcolor{black}{(b)}. Specifically, the number of kernels in each convolutional layer of the side block is set to be the same as that of the output feature channels from the corresponding main block.
For adapter design, if the spatial size of the original output feature maps from the main block is smaller (e.g., due to pooling) 
than that of the side block output, the adapter $g(\cdot)$ is designed as a max pooling operator (e.g., max over local regions of size $2 \times 2$ elements with step size 2) to reduce the spatial size of side block output such that two outputs have the same spatial size and the same number of feature channels. Otherwise,  the adapter $g(\cdot)$ is simply an identity function.
{For instance, when the baseline iCaRL~\cite{rebuffi2017icarl} with the CNN backbone ResNet is adopted as the main branch for the proposed G2B framework, 
the adapter $g(\cdot)$ for the first residual block is a max pooling operator, while the adapters for the remaining blocks are simply identity functions.}

\textcolor{black}{When a Transformer backbone is used in the main branch, the number of kernels in each convolutional layer of the side block is set to the length of the output vector for each token in the corresponding main block.}
The adapter $g(\cdot)$ is designed as a simple convolutional layer followed by a reshape operator such that the transformed output of the side branch by the adapter has the same dimension as that of the corresponding main block output, as detailed in Figure~\ref{fig:framework}\textcolor{black}{(c)}. 
\textcolor{black}{Specifically,}
suppose the dimension of the main block output is $[D, P]$, where $P$ is the number of tokens (with each token corresponding to one local image patch) and $D$ is the length of vector output for each token, and the dimension of the corresponding side block output is $[D, S, S]$, where $S$ is both width and height of the output feature maps. Then the adaptor $g(\cdot)$ contains $D$ convolutional kernels with kernel size $[S/\sqrt{P}, S/\sqrt{P}]$ and convolution step size $S/\sqrt{P}$ ($S/\sqrt{P}$ being an integer number in practice). The dimension (i.e., $[D, \sqrt{P}, \sqrt{P}]$) of the adapter output is reshaped to $[D, P]$ by simply re-ordering the $D$-dimensional feature vectors across all spatial locations in the adapter output such that each feature vector in the reshaped adapter output matches one token's output vector in the corresponding main block. 
{Taking DyTox~\cite{DyTox} as an example. Its ViT backbone network serves as the main branch of the G2B framework. Each Transformer block $\mathbf{F}_i$ has a corresponding convolutional block $\mathbf{M}_i$ that generates a corresponding soft mask through the adapter $g(\cdot)$. 

The training of the proposed G2B framework is straightforward. Its main and side branches are jointly optimized in an end-to-end manner. The losses used in the vanilla method can be directly inherited by G2B. No extra loss is introduced. The optimization settings, e.g., learning rate and optimizer, of G2B are kept consistent with those in the vanilla method.}

\begin{table}[t]
\centering
\footnotesize
\caption{Representative CIL methods with their main strategies and model backbones. `RH', `RG' and `DE' correspond to the CL strategies, `Rehearsal-based', `Regularization-based' and `Dynamic Expansion', respectively.}
    \vspace{-0.2cm}
    \resizebox{\linewidth}{!}{\begin{tabular}{c|cccccc}
\hline
Method & iCaRL\cite{rebuffi2017icarl} & BiC\cite{wu2019bias_correction} & WA\cite{zhao2020maintaining} & PODNet\cite{douillard2020podnet} & DER\cite{yan2021der} & DyTox\cite{DyTox} \\ \hline
Strategy & RH & RG & RG & RG & DE & DE \\ \hline
Backbone & CNN & CNN & CNN & CNN & CNN & ViT \\ \hline
\end{tabular}}
\label{tab:CIL_vanilla}
\end{table}

\section{Experiments}

\noindent\textbf{Datasets and protocols.} CIFAR-100~\cite{krizhevsky2009cifar100}, ImageNet-100 and ImageNet-1000~\cite{deng2009imagenet} are widely used benchmarks for CIL.
CIFAR-100 consists of 32$\times$32 RGB images of 100 object categories. It contains 50,000 training samples with 500 images per class and 10,000 testing images with 100 samples per class.
ImageNet-1000 is a large-scale vision dataset of 1,000 object categories and includes approximately 1.2 million RGB images for training and 50,000 images for validation. ImageNet-100 is a subset of the 100 classes in ImageNet-1000 and the same 100 classes as in
DER and DyTox are used.
The CIL protocols in DyTox~\cite{DyTox} and DER~\cite{yan2021der} are followed for fair comparisons.
On CIFAR-100, the models are gradually trained with 10, 5 or 2 new classes per round (task), resulting in three continual learning procedures: 10 rounds, 20 rounds or 50 rounds of continual learning, respectively.
The fixed rehearsal memory size is 2,000 exemplars.
The CIL procedure on ImageNet includes 10 rounds of continual learning.
For ImageNet-100, each round provides the training samples of 10 new classes with a fixed memory size of 2,000.
For ImageNet-1000, each round provides the training samples of 100 new classes with a fixed memory size of 20,000. 
After each learning round, the classification accuracy of all the classes learned so far in the test set is obtained.
After finishing the last round of continual learning, both the averaged classification accuracy over all rounds (denoted as Avg) and the last-round accuracy (denoted as Last) are reported.

\begin{table}[t]
\centering
\caption{CIL Results (\%) on CIFAR-100.}
\vspace{-0.2cm}
\resizebox{0.50\textwidth}{!}{%
\begin{tabular}{@{}c|c|ll|ll|ll}
\hline
 & & \multicolumn{2}{c|
}{10 rounds} & \multicolumn{2}{c|}{20 rounds} & \multicolumn{2}{c}{50 rounds}\\
\textbf{Method} &\begin{tabular}[c]{@{}c@{}}\textbf{Strategy \&}\\ \textbf{Backbone}\end{tabular}  & \textbf{Avg} & \textbf{Last} & \textbf{Avg} & \textbf{Last} & \textbf{Avg} & \textbf{Last}\\
\hline
\multirow{2}{*}{Joint Learning}  & - \& CNN & - & 80.41 & - & 81.49 & - & 81.74\\
                     & - \& ViT & - & 76.12 & - & 76.12  & - & 76.12\\
\hline
iCaRL  & \multirow{2}{*}{RH \& CNN} & 65.71 & 52.97 & 61.20 & 43.75 & 56.08 & 36.62\\
G2B(iCaRL)   & & 68.57\padd{2.86} & 55.75\padd{2.78} & 65.12\padd{3.92} & 46.81\padd{3.06} & 58.21\padd{2.13} & 38.78\padd{2.16}\\ \hline
BiC  & \multirow{2}{*}{RG \& CNN} & 68.83 & 52.71 & 66.48 & 47.02 & 62.09 & 41.04\\
G2B(BiC) & & 71.11\padd{2.28} & 54.63\padd{1.92} & 68.21\padd{1.73} & 49.43\padd{2.41} & 64.13\padd{2.04} & 42.32\padd{1.28}\\ \hline
WA   & \multirow{2}{*}{RG \& CNN} & 70.10 & 53.68 & 67.33 & 47.31 & 64.32 & 42.14\\
G2B(WA) & & 73.45\padd{3.35} & 58.58\padd{4.90} & 69.81\padd{2.48} & 50.32\padd{3.01} & 65.81\padd{1.49} & 44.35\padd{2.21}\\ \hline
PODNet  & \multirow{2}{*}{RG \& CNN} & 61.21 & 42.10 & 53.97 & 35.02 & 51.19 & 32.99\\
G2B(PODNet) & & 65.05\padd{3.84} & 44.07\padd{1.97} & 56.31\padd{2.34} & 36.19\padd{1.17} & 53.12\padd{1.93} & 35.12\padd{2.13}\\ \hline
DER  & \multirow{2}{*}{DE \& CNN} & 76.08 & 66.59 & 74.09 & 62.48 & 72.41 & 59.08\\ 
G2B(DER) & & 77.26\padd{1.18} & 68.36\padd{1.77} & 74.95\padd{0.86} & 63.15\padd{0.67} & 73.81\padd{1.40} & 59.86\padd{0.78}\\ \hline
DyTox  & \multirow{2}{*}{DE \& ViT} & 73.66 & 60.67 & 72.27 & 56.32 & 70.20 & 52.34\\
G2B(DyTox) & & 76.32\padd{2.66} & 64.32\padd{3.65}  & 75.30\padd{3.03} & 61.04\padd{4.72} & 73.01\padd{2.81} & 55.88\padd{3.54}\\ \hline
\end{tabular}
}
\label{tab:cifar100}
\end{table}

\begin{figure}[t]
	\centering
	\includegraphics[width=0.46\textwidth]{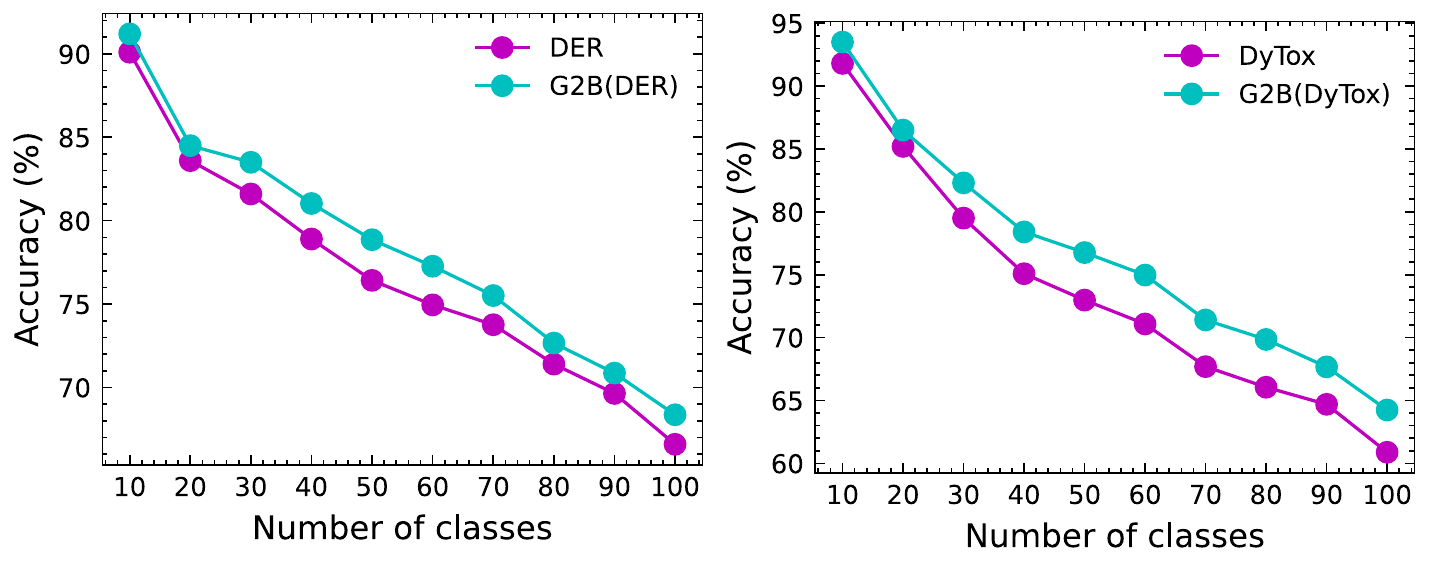}
 \vspace{-0.2cm}
	\caption{
 Comparisons between the G2B framework and corresponding methods, DER (left) and DyTox (right), on CIFAR-100 under 10 rounds of continual learning.
 }
	\label{fig:cifar100}
\vspace{-0.2cm}
\end{figure}

\noindent\textbf{Vanilla CIL methods.}
Our G2B framework is proposed to further enhance the performance of existing CL methods. 
As shown in Table~\ref{tab:CIL_vanilla}, six representative CIL methods are chosen to compare and combine with our G2B.
ResNet18~\cite{he2016deep} is the default CNN backbone that existing CIL methods have widely adopted, and the ViT backbone is adopted from DyTox~\cite{DyTox}. 
Training a G2B model can be done with the vanilla optimization setting applied directly.

\begin{table}[t]
\footnotesize
		\caption{Ten rounds of CIL results (\%) on ImageNet-100.}
    \vspace{-0.2cm}
    \resizebox{\linewidth}{!}{
			\begin{tabular}{c|llll}
			\toprule[0.3mm]
			 & \multicolumn{2}{c}{\textbf{Top-1}\ \ \ \ \ \ \ \ \ \ \ \ }  & \multicolumn{2}{c}{\textbf{Top-5}\ \ \ \ \ \ \ \ \ \ \ \ }  \\
			\textbf{Method}   & \textbf{Avg}           & \textbf{Last}        & \textbf{Avg}     & \textbf{Last}      \\
			\hline
			iCaRL & 68.22 & 53.02 & 87.68 & \textcolor{black}{75.64} \\ 
			G2B(iCaRL) & 69.12\padd{0.90} & 54.14\padd{1.12} & 87.98\padd{0.30} & \textcolor{black}{76.64}\padd{1.00} \\ \hline
			BiC  &65.39 & 48.76 & 85.31 & 69.62 \\
			G2B(BiC) & 68.13\padd{2.74} & 51.48\padd{2.72} &87.29\padd{1.98} & 72.54\padd{2.92} \\ \hline
			WA  & 70.28 & 57.42 & 89.30 & 80.24  \\
			G2B(WA) & 73.56\padd{3.28} & 61.64\padd{4.22} & 91.31\padd{2.01} & 84.14\padd{3.90} \\ \hline
                DER   & 77.05 & 66.28  & 93.42 & 87.66  \\ 
                G2B(DER) & 79.22\padd{2.17} & 69.21\padd{2.93} & 93.87\padd{0.45} & 87.99\padd{0.33} \\ \hline
			DyTox  & 77.15 & 69.10 & 92.04 & 87.98 \\
			G2B(DyTox) & 80.91\padd{3.76} & 69.94\padd{0.84} & 94.58\padd{2.54} & 90.38\padd{2.40} \\
			\hline
			\end{tabular}
   }
	\label{tab:imagenet_100}
\end{table}

\begin{table}[t]
\footnotesize
		\caption{Ten rounds of CIL results (\%) on ImageNet-1000.}
    	\vspace{-0.2cm}
    \resizebox{\linewidth}{!}{
			\begin{tabular}{c|llll}
			\toprule[0.3mm]

			 & \multicolumn{2}{c}{\textbf{Top-1}\ \ \ \ \ \ \ \ \ \ \ \ \ \ \ }  & \multicolumn{2}{c}{\textbf{Top-5}\ \ \ \ \ \ \ \ \ \ \ \ \ \ \ }  \\
			\textbf{Method}   & \textbf{Avg}           & \textbf{Last}        & \textbf{Avg}     & \textbf{Last}      \\
			\hline
			WA  & 65.67 & 55.60 & 86.60 & 81.10  \\
			G2B(WA) & 67.17\padd{1.50} & 58.92\padd{3.32} & 88.61\padd{2.01} & 83.12\padd{2.02} \\ \hline
                DER   & 68.84 & 60.16 & 88.17 & 82.86  \\ 
                G2B(DER) & 69.56\padd{0.72} & 62.13\padd{1.97} & 89.19\padd{1.02} & 83.24\padd{0.38} \\ \hline
			DyTox  & 71.29 & 63.34 & 88.59 & 84.49 \\
			G2B(DyTox) & 73.16\padd{1.87} & 63.87\padd{0.53} & 89.37\padd{0.78} &86.62\padd{2.13} \\
			\hline
			\end{tabular}
   }
	\label{tab:imagenet_1000}
\end{table}

\subsection{Main Results}

The proposed G2B framework along with the corresponding CIL methods are first evaluated on CIFAR-100 dataset, as shown in Table~\ref{tab:cifar100}.
Joint learning indicates the backbone networks are optimized with all training samples and depicts the upper-bound performance of continual learning.
Combining our G2B framework with the vanilla CIL methods results in clear and consistent improvements.
Specifically, it boosts the performance of the SOTA methods, DER and DyTox, to new peaks.
Combining DER with G2B brings consistent improvements, usually more than $1\%$, on all criteria.
DyTox also substantially benefits from the G2B framework, with significant improvements observed, more than $2.6 \%$.
The effectiveness of G2B in defying catastrophic forgetting can be further demonstrated in Figure~\ref{fig:cifar100} where the performance of DER and DyTox is evaluated at the end of each continual learning round.
The initial results of the vanilla method and with G2B are similar.
As continual learning progresses, the G2B framework consistently achieves better results than its counterpart.

Our G2B framework is also evaluated on the two ImageNet datasets. 
ImageNet-100 results are shown in Table~\ref{tab:imagenet_100} and the ImageNet-1000 ones are summarized in Table~\ref{tab:imagenet_1000}. The latter one could be a more challenging CIL setting with more categories and data samples to be handled at each learning round.
However, both tables show that combining the vanilla CIL methods with our G2B framework consistently increases their performance with clear margins.
In addition, new SOTA results are achieved by G2B(DER) and G2B(DyTox).

\begin{table}[t]
\centering
\footnotesize
\caption{The effect of G2B side branch network, with 10-round CIL results of CIFAR-100 and ImageNet-100 reported.
$\#$P: the number of (million) parameters of a model. `-': not applicable because of only four blocks for the main branch.}
\vspace{-0.2cm}
\resizebox{\linewidth}{!}{
\begin{tabular}{c|ccccc|c|cc|cc}
\hline
\multirow{2}{*}{Method} & \multicolumn{5}{c| }{Side Branch Blocks}  & &\multicolumn{2}{c| }{CIFAR-100} & \multicolumn{2}{c}{ImageNet-100} \\
                         &  1    & 2   &  3    &  4  & 5 & $\#$P   & Avg          & Last          & Avg            & Last           \\ \hline

\multirow{5}{*}{WA}    &  &    &     &  & -  & 11.2 & 70.16           &     53.19     &      70.28          &   57.42             \\ 
                    & \checkmark &    &     &  & -  & 11.3 &  70.52           &     53.21     &   72.17             &     57.00           \\
                    & \checkmark & \checkmark   &      &     & -            &  11.5         &   70.63           &     53.96        &   72.91             &   57.60             \\
                    & \checkmark & \checkmark   & \checkmark     &  & -  & 12.0 & 71.47           &   55.09          &   72.52           &   58.28           \\
                    & \checkmark & \checkmark   & \checkmark     & \checkmark   & -  & 13.5 & 73.45           &    58.58              &   73.56             &  61.64              \\ \hline

\multirow{6}{*}{DyTox} &  &    &     &    &      & 11.0 &  73.66           &    60.67           &   77.15          & 69.10        \\
                        & \checkmark &    &     &    &   & 11.9 &   73.82           &    60.52           &    77.61            &    68.98            \\ 
                        & \checkmark & \checkmark   &   &   &  & 12.5 &   74.15           &   61.56            &   78.67             & 69.16               \\
                        & \checkmark & \checkmark   & \checkmark     &   &  & 13.1  &   74.58    &  62.15    &     80.14       &   69.48          \\
                        & \checkmark & \checkmark   & \checkmark     & \checkmark   &       & 13.7 &  75.62   &    63.18   &   80.58    &  69.89              \\
                        & \checkmark & \checkmark   & \checkmark     & \checkmark   &   \checkmark     & 14.3 & 76.32     &    64.32     &    80.91            &     69.94           \\\hline
\end{tabular}}
\label{tab:cnn_vit_blocks_paras}
\end{table}

\begin{table}[t]
	\centering
 \footnotesize
	\caption{
 Impact of model size.
 Ten-round of CIL results on CIFAR-100 are reported.
 ResNet18 is the backbone network.
 }
	\vspace{-0.2cm}
        \begin{tabular}{c|c|c|c|c}
        \hline
        Methods & $\#$P & Avg & Last &$F_{10}$ ($\downarrow$)  \\\hline
        \multirow{3}{*}{WA} &{11.2}M & 70.16   &53.19  &0.3437 \\ 
           &14.5M & 70.69   & 54.11  &0.3408\\
           &21.5M & 71.53   & 54.29  &0.3559\\\hline
        \multirow{3}{*}{G2B(WA)} & 7.81M & 73.14   &57.81  &0.2430\\
        & 10.6M & 73.93   & 58.92 &0.2340\\ 
          &{13.5}M  & 73.45   & 58.58  &0.2659\\ \hline
        \end{tabular}
	\label{tab:paras_FI}
\end{table}

\begin{table}[!t]
	\centering
  \footnotesize
	\caption{Comparing G2B with AAnets, with 10/20-round CIL results ($\%$) on CIFAR-100. ResNet18 is the backbone.}
  	\vspace{-0.2cm}
\begin{tabular}{c|cccc}
\hline
                        & Memory & FLOPs  & 10    & 20     \\\hline
        iCaRL       & 11.38MB & 557.4M & 65.71 &61.20 \\  
        AANets(iCaRL)   & 21.73MB & 1114.8M  &62.49 &59.35 \\ 
        G2B(iCaRL)    & 19.28MB   &704.7M & 68.57 &65.12 \\ \hline
        \end{tabular}
	\label{tab:cmp_aanet}
\end{table}

\subsection{Detailed Analysis}

\textbf{Effect of the side branch network.} 
As illustrated in Figure~\ref{fig:framework}, G2B contains two parallel networks.
To evaluate the contributions of the newly introduced side branch, the side branch blocks are gradually removed, and the performance is reported in Table~\ref{tab:cnn_vit_blocks_paras}.
The side branch network with two blocks consistently brings noticeable improvement over the vanilla method.
With more blocks used in the side branch, better results are obtained.
The complete G2B, where each main block has the corresponding side block, achieves the best results.

{ \noindent\textbf{Impact of model size.}
The performance of models under varied parameter sizes are compared in Table~\ref{tab:paras_FI}.
Strengthening the vanilla method, WA, from $11.2$M to $21.5$M parameters brings only marginal improvements.
The proposed G2B is insensitive to smaller model sizes and consistently better than WA.
To demonstrate the superior stability of our G2B framework, the forgetting measure ($F$)~\cite{RiemannianWalk} is used. We report $F_{10}$ to measure this forgetting of the final ($10$th) round model. 
As shown in the last column of Table~\ref{tab:paras_FI}, the $F$ values of G2B are much lower than those of WA.
}

{ \noindent\textbf{Comparing two-branch frameworks.}
Both our G2B and AANet~\cite{AANets} are two-branch network architectures.
As shown in Table~\ref{tab:cmp_aanet}, AANets requires more memory and computation resources than the vanilla iCaRL and our G2B method but achieves the worst performance.

\section{Conclusion}
We propose a simple yet effective generalizable two-branch (G2B) framework for continual learning.
With the lightweight convolutional network as the side branch, the outputs in the main branch are modulated for relative sparsity and thus alleviating catastrophic forgetting.
G2B is compatible with different continual learning strategies based on either CNN or ViT. Their combinations are straightforward and the optimization is painless.
Comprehensive evaluations on multiple image classification tasks demonstrate the state-of-the-art continual learning performance of the G2B framework.
We expect that G2B 
can be extended to continual learning of object detection and semantic segmentation tasks.

\noindent\textbf{Acknowledgement.}\quad This research is supported by the National Natural Science Foundation for Young Scientists of China (No. 62106289).

\bibliographystyle{IEEEbib}
\bibliography{refs}

\end{document}